\documentclass{article}

\usepackage{arxiv}

\usepackage[utf8]{inputenc} 
\usepackage[T1]{fontenc}    
\usepackage{hyperref}       
\usepackage{url}            
\usepackage{booktabs}       
\usepackage{amsfonts}       
\usepackage{nicefrac}       
\usepackage{microtype}      
\usepackage{lipsum}
\usepackage{graphicx}
\graphicspath{ {./images/} }

\usepackage{amsmath}

\usepackage{multirow}
\usepackage{subcaption}  
\usepackage{lipsum}  
\usepackage{float}   
\usepackage{stfloats} 
\usepackage{algorithm}
\usepackage{algorithmic}

\title{Spiking Neural Networks with Temporal Attention-Guided Adaptive Fusion for imbalanced Multi-modal Learning}

\author{
\textbf{Jiangrong Shen \dag \ddag \S \ss \quad
Yulin Xie \dag \quad  Qi Xu \P   \quad Gang Pan \S \quad Huajin Tang  \S \quad  Badong Chen \ddag \ss} \\
\dag Faculty of Electronic and Information Engineering,
Xi’an Jiaotong University \\
\ddag Institute of Artificial Intelligence and Robotics, Xi’an Jiaotong University \\
\S State Key Lab of Brain-Machine Intelligence, Zhejiang University \\
\P School of Computer Science, Dalian University of Technology \\
\ss National Key Lab of Human-Machine Hybrid Augmented Intelligence, Xi’an Jiaotong University
}

\begin{document}
\maketitle
\begin{abstract}

Multimodal spiking neural networks (SNNs) hold significant potential for energy-efficient sensory processing but face critical challenges in modality imbalance and temporal misalignment. Current approaches suffer from uncoordinated convergence speeds across modalities and static fusion mechanisms that ignore time-varying cross-modal interactions. We propose the temporal attention-guided adaptive fusion framework for multimodal SNNs with two synergistic innovations: 1) The Temporal Attention-guided Adaptive Fusion (TAAF) module that dynamically assigns importance scores to fused spiking features at each timestep, enabling hierarchical integration of temporally heterogeneous spike-based features; 2) The temporal adaptive balanced fusion loss that modulates learning rates per modality based on the above attention scores, preventing dominant modalities from monopolizing optimization. The proposed framework implements adaptive fusion, especially in the temporal dimension, and alleviates the modality imbalance during multimodal learning, mimicking cortical multisensory integration principles. Evaluations on CREMA-D, AVE, and EAD datasets demonstrate state-of-the-art performance (77.55\%, 70.65\% and 97.5\%accuracy, respectively) with energy efficiency. The system resolves temporal misalignment through learnable time-warping operations and faster modality convergence coordination than baseline SNNs. This work establishes a new paradigm for temporally coherent multimodal learning in neuromorphic systems, bridging the gap between biological sensory processing and efficient machine intelligence.

\end{abstract}





\section{Introduction}
\label{intro}

The human brain achieves robust environmental perception through precisely coordinated multisensory integration, where neural circuits in posterior cortical regions dynamically synthesize inputs from distributed sensory pathways \cite{glasser2016multi}. Particularly, rhythmic cortical activity has been shown to support motion-related multisensory integration \cite{shen2025cortical}, with evidence suggesting that hierarchical temporal coding mechanisms orchestrate the processing of naturalistic stimuli across multiple timescales. This mechanism emphasizes the criticality of temporal dynamics among sensory modalities during cross-modal information fusion.
In artificial intelligence systems, multimodal learning frameworks that combine multiple different modalities have demonstrated remarkable advantages over unimodal counterparts \cite{jiang2023cmci,liu2022event}. These benefits stem from cross-modal complementarity—where weaknesses in one modality (e.g., visual occlusion) can be compensated for by others (e.g., auditory localization) \cite{peng_balanced_2022}. Advanced fusion techniques like attention-based feature recombination \cite{ye2024mplug} further enhance model efficiency. However, conventional artificial neural networks (ANNs) suffer from the limitation of energy costs due to continuous-valued computations and the biologically implausible integration mechanisms that neglect the temporal coordination principles observed in neural systems.


Spiking neural networks (SNNs)~\cite{maass_networks_1997, subbulakshmi2021biomimetic, spikejelly, xu2023enhancing, shen2024efficient} mimic the spatiotemporal dynamics of biological neurons, providing a more biologically plausible approach to modeling neural processes compared to traditional ANNs. Unlike conventional ANN-based models, SNNs process information using discrete spikes, which are more reflective of how real neurons communicate. The event-driven nature of SNNs, where neurons only fire in response to meaningful stimuli, allows them to handle spatiotemporal data efficiently while consuming significantly less energy \cite{chen2022state, zhou2023computational, shi2023towards, xu2024reversing, shen2025improving}. This event-based computation also results in lower power consumption compared to traditional ANN models, which process data in a continuous manner.
SNNs also exhibit richer spatiotemporal dynamics, enabling to capture complex temporal patterns and interactions across different timescales. This dynamic feature is particularly advantageous in tasks involving multisensory integration, where the timing and synchronization of inputs across modalities are crucial. Therefore, the ability of SNNs to leverage temporal dynamics enhances their capacity to model different temporal structures of stimuli with different modalities, which is essential for effective multisensory processing.

Despite these advantages, current SNN-based multi-modal models still face challenges related to modality imbalance, which arises from uncoordinated convergence and temporal inconsistencies between modalities. Moreover, the temporal dynamics among different modalities that hierarchically organize multisensory integration are often overlooked in existing SNN architectures, preventing the full exploitation of the rich spatiotemporal information inherent in multi-modal learning. 
In detail, the challenge of modality imbalance in spiking multimodal learning arises from the inherent heterogeneity of sensory modalities and the temporal misalignment in their spiking dynamics. Specifically, different modalities exhibit distinct temporal processing requirements: static visual inputs demand shorter timesteps for efficient feature extraction, while dynamic visual streams and auditory signals necessitate progressively longer timesteps to capture their temporal intricacies. Conventional approaches that enforce uniform timestep alignment across modalities inevitably lead to information redundancy or loss, as the fixed temporal resolution fails to adapt to modality-specific spiking patterns. Moreover,
while assigning modality-specific timesteps to SNN branches can partially address temporal mismatches in multimodal learning, the fusion of these temporally heterogeneous representations remains fundamentally constrained by static aggregation mechanisms. These static aggregation approaches, which unify multimodal spike trains through simplistic temporal averaging or fixed-weight summation among different time steps, often fail to capture the time-varying contributions of different modalities to joint decision-making. 
Compounding this issue, the varying convergence speeds of unimodal spiking neural pathways result in dominant modalities monopolizing the optimization process, leaving others undertrained due to insufficient supervisory feedback.

To address these intertwined challenges, we propose a temporal attention-guided adaptive fusion framework for multimodal SNNs that harmonizes multimodal spiking dynamics through two synergistic mechanisms. First, temporal attention scores dynamically recalibrate loss weights across timesteps, prioritizing crucial temporal features unique to each modality. Second, attention-guided gradient modulation balances convergence speeds by adaptively adjusting learning rates based on inter-modal training progress disparities, preventing premature dominance of faster-converging modalities. This biologically inspired approach resolves temporal misalignment and enhances energy efficiency. By mirroring the brain’s ability to prioritize temporally salient inputs, our framework establishes a new paradigm for energy-efficient, temporally coherent, balanced multimodal integration in SNNs, where temporally distinct neural features selectively amplify or suppress modality-specific inputs based on their temporal congruence and task relevance. 

The contributions are summarized as follows:
\begin{itemize}
    \item We propose the temporal attention-guided adaptive fusion (TAAF) module for multimodal SNNs. It assigns temporal importance scores at every timestep to modulate both feature fusion, which improves the temporal hierarchy of sensory integration.
    \item The temporal importance scores from TAAF module are integrated into temporal adaptive balanced fusion loss function to enable dynamic modality-specific convergence speed adjustment for each modality branch at individual timesteps. 
    \item The experiments are conducted on three datasets with the comparison experiments and ablation study. Our model achieves the accuracy of 77.55\%, 70.65\%, and 97.5\% on CREMA-D, AVE and EAD datasets, alleviating the modality imbalance by enhancing the temporal dynamics of SNNs.
\end{itemize}

\section{Related Works}

In this section, we discuss the recent multimodal learning studies with SNNs according to the different applications. Then we summarize the typical methods for the imbalance problem in multimodal learning.

\subsection{Multimodal learning in SNNs}

The exploration of multimodal learning in SNNs has gained momentum with recent advances in neuromorphic computing. We summarize these studies according to different categories of applications.

As to multimodal learning with visual and auditory modalities, the unsupervised learning of multimodal SNNs that integrates two modalities (image and audio) is explored in \cite{rathi2018stdp}, where two unimodal ensembles are linked via cross-modal connections. The excitatory connections within the unimodal ensemble and the cross-modal connections are trained using a power-law weight-dependent spike-timing-dependent plasticity (STDP) learning rule. Experimental results demonstrate that the multimodal network captures features from both modalities and enhances classification accuracy compared to unimodal architectures, even when one modality is affected by noise.

As to the Audio-Visual Classification (AVC) task, it is explored in \cite{liutowards}, where the spike activations of audiovisual signals are synchronized and coordinated within a single neuron by introducing a multimodal Leaky Integrate-and-Fire (MLIF) neuron. The event-based multimodal SNN is introduced in \cite{liu2022event}, consisting of visual and auditory unimodal subnetworks, along with an attention-based cross-modal subnetwork for fusion.

The combination of the visual modality branch (Neuromorphic-MNIST [N-MNIST]) and the auditory modality branch (Spiking Heidelberg Digits [SHD]) is analyzed in \cite{bjorndahl2024digit} to assess the performance of SNNs in digit classification. Additionally, the cross-modality current integration (CMCI) for multimodal SNNs is proposed in \cite{jiang2023cmci} to perform visual, auditory, and olfactory fusion recognition tasks. The spiking multimodal transformer that integrates SNNs and Transformers is also proposed for multimodal audiovisual classification \cite{guo2023transformer}.

As to emotion recognition tasks with multimodal SNNs, the evolving architecture of multimodal SNNs based on the NeuCube framework is utilized in \cite{tan2020fusionsense}. The multimodal data includes facial expressions along with physiological signals such as ECG, skin temperature, skin conductance, respiration signals, mouth length, and pupil size. The Sliding Parallel Spiking Convolutional Vision Transformer (SPSNCVT) is designed for robust and efficient multimodal emotion recognition \cite{chen2025enhancing}. This model integrates facial expressions, voice, and text, using aligned heatmap features and multiscale wavelet transforms for precise feature extraction.

The spiking multimodal interactive label-guided enhancement network for emotion recognition is proposed in \cite{guo2024smile}, incorporating modality-interactive exploration and label-modality matching modules to capture multimodal interaction and label-modality dependence.

As to multimodal brain data learning, considering the heterogeneous temporal and spatial characteristics of various brain data modalities, \cite{wysoski2010brain} design an unsupervised learning algorithm for fusing temporal, spatial, and orientation information within an SNN framework. This could potentially be used to understand and model predictive tasks based on multimodal brain data.

As to joint learning between event and frame modalities, the fusion method based on cross-modality attention (CMA) is introduced in \cite{zhou2024enhancing} to effectively capture the unique benefits of each modality. The CMA learns both temporal and spatial attention scores from the spatiotemporal features of event and frame modalities and allocates these scores across modalities to enhance synergy.

There are also other applications of multimodal SNNs, such as early bearing fault diagnosis \cite{xu2025multi}, skeleton-based action recognition \cite{zheng2024mk, zheng2025snn}, person authentication \cite{wysoski2010brain}, and dynamic obstacle avoidance \cite{wang2023event}. Additionally, studies have been conducted to implement multimodal-fused SNNs on hardware platforms. For example, multimodal-fused spiking neural arrays have been developed to enhance tactile pattern recognition by heterogeneously integrating a pressure sensor to process pressure and a NbOx-based memristor to sense temperature \cite{zhu2022heterogeneously}.

Moreover, there are hybrid ANN-SNN models for multimodal learning. The SSTFormer in \cite{wang2023sstformer} integrates RGB frames and event streams simultaneously to recognize RGB-frame-event patterns. The SSTFormer includes a memory-supported Transformer network for RGB frame encoding, an SNN for encoding raw event streams, a multimodal bottleneck fusion module for RGB-Event feature aggregation, and a prediction head. To enhance feature extraction for object tracking, the multimodal hybrid tracker (MMHT) is introduced in \cite{sun2024reliable}, utilizing frame-event-based data for reliable single-object tracking. The hybrid backbone of MMHT combines an ANN and an SNN to extract dominant features from both frame and event visual modalities, followed by a unified encoder and transformer-based module to enhance features across different domains.
The embedded multimodal monocular depth estimation framework, using a hybrid SNN and ANN architecture, is implemented in \cite{tumpa2024snn}. It employs a custom accelerator, TransPIM, for efficient transformer deployment, enabling real-time depth estimation on embedded systems. The TempSimNet, which integrates long-short-term memory (LSTM) networks with SNNs, is introduced in \cite{li2024hybrid} for effective temporal feature extraction and dynamic processing, improving the model's ability to recognize unseen classes in multimodal datasets. In this framework, LSTM excels at extracting long-term temporal dependencies, while the SNN processes these features with high temporal precision through spike-based encoding.

Despite these advancements, most existing multimodal SNN systems overlook the issue of modality imbalance and still face challenges in handling asynchronous modality inputs and temporal-scale mismatches. Current methods mainly focus on pairwise modality combinations (e.g., vision-audio), leaving temporal multimodal interactions in SNNs underexplored.






\subsection{Multimodal learning methods for imbalanced problem}

Multi-modal learning with ANNs \cite{huang2021makes} has demonstrated impressive performance across various research domains, such as brain data analysis \cite{tang2023explainable, van2018learning}, video understanding \cite{li2024mvbench}, time series \cite{liu2024time}, motion measurement \cite{huang2025ridge, huang20253d, zhao2025full, liang2025accurate}, vision navigation \cite{liang2025high, yu2021new}and large language models \cite{ye2024mplug}. In this paper, we primarily address the issue where multi-modal models tend to underperform relative to their original unimodal counterparts \cite{wang_what_2020}, and where the unimodal branches fail to surpass the performance of individual unimodal models \cite{peng_balanced_2022}.

Recently, a variety of strategies have been proposed to enhance the multi-modal learning of ANNs by achieving better balance across different modalities. One notable approach is minimizing the overfitting-to-generalization ratio (OGR) through an optimal combination of multiple supervision signals. This strategy jointly optimizes the overfitting and generalization rates for the various modalities \cite{wang_what_2020}. The gradient-blending technique computes an optimal fusion of modalities based on their overfitting behavior, thereby improving the end-to-end training of ensemble models in multi-modal learning.
Another important contribution comes from \cite{sun_learning_2021}, where the learning processes of unimodal and multimodal networks are decoupled by dynamically adjusting the learning rates for different modalities. Using an adaptive tracking factor (ATF) that updates the learning rate for each modality in real-time, this method employs adaptive convergent equalization (ACE) and bilevel directional optimization (BDO) to regulate and update the ATF, preventing the suboptimal unimodal representations caused by overfitting or underfitting.
Further, \cite{wu_characterizing_2022} introduces a conditional utilization rate to estimate a model's reliance on each modality by computing the gain in accuracy when the model has access to a modality in addition to another one. The balanced multi-modal learning approach, which leverages conditional learning speeds, is designed to guide the model in learning from previously underutilized modalities.
Additionally, \cite{peng_balanced_2022} presents on-the-fly gradient modulation, a method that adaptively controls the optimization of each modality by monitoring the discrepancy in their contributions towards the learning objective. To prevent potential generalization drops caused by gradient modulation, dynamic Gaussian noise is introduced as a safeguard.

These methods share a common objective: optimizing the multi-modal training process by addressing the issue of under-optimized unimodal representations caused by the dominance of other modalities in certain contexts. These studies offer valuable insights into solving the imbalance problem in SNNs.
To capture the spatiotemporal dependencies between modalities and address the semantic mismatch problem, \cite{he2025enhancing} designs a cross-modal complementary spatiotemporal spiking attention module and semantic alignment optimization. This module consists of two components: Temporal Complementary Spiking Attention (TCSA) and Spatial Complementary Spiking Attention (SCSA).
TCSA is designed to extract complementary information along the temporal dimension, capturing temporal correlations across different modalities and facilitating the integration of temporal data. On the other hand, SCSA focuses on spatial fusion across modalities by modeling correlations between spatial locations. However, the aforementioned work does not fully explore the temporal dynamics at each time step in SNNs, which limits the potential for advancing multimodal learning within these networks.

\section{Method}

\begin{figure*}[t]
  \centering
  \includegraphics[width=1\textwidth]{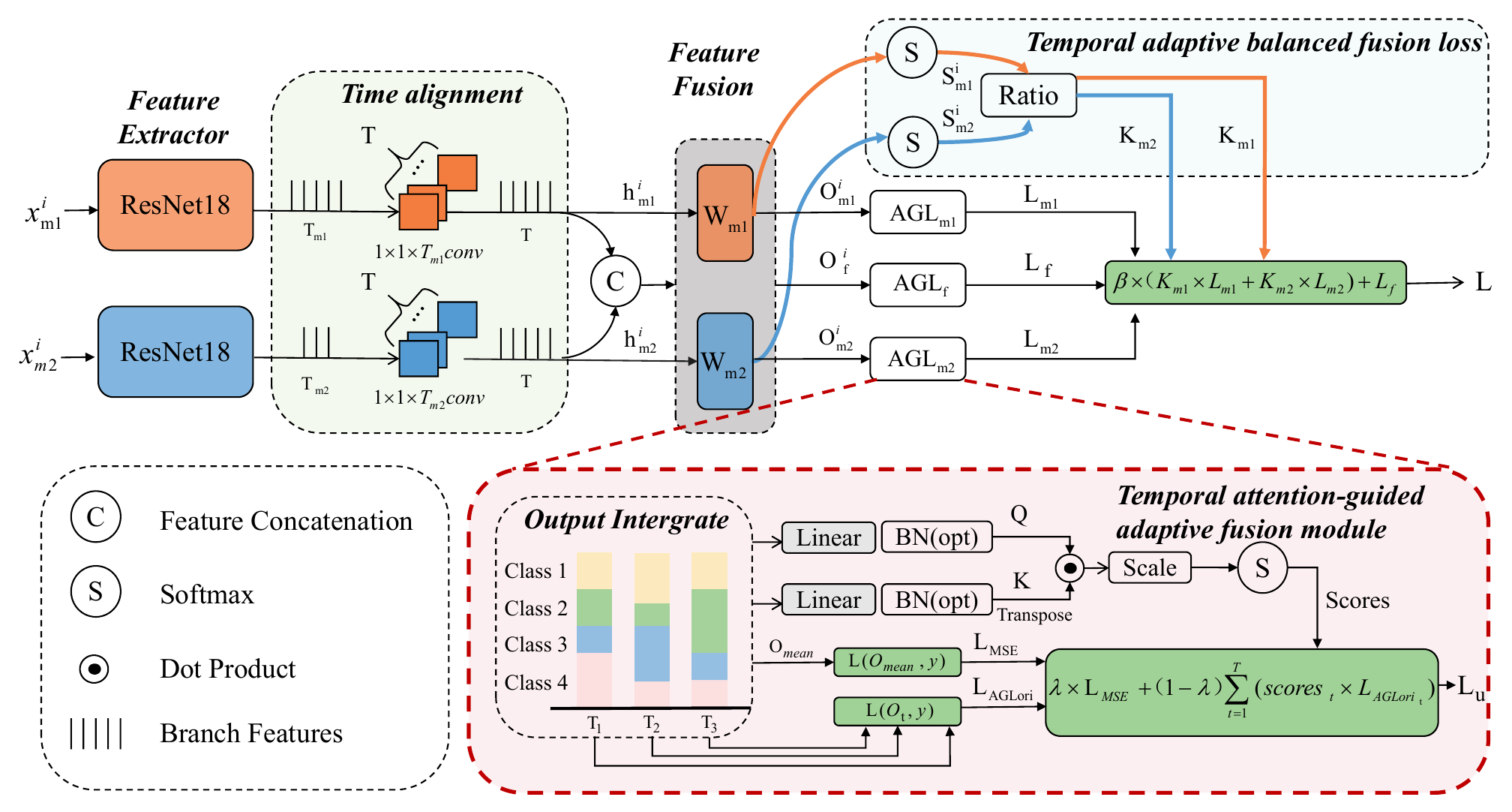}
  \caption{Architecture of the proposed attention-guided multimodal SNN.  The temporal attention-guided adaptive fusion (TAAF) module generates temporal importance scores at every timestep that modulate both feature fusion and loss calculation. The temporal importance scores are then integrated into the temporal adaptive balanced fusion loss to enable dynamic adjustment for each modality branch in individual timesteps. }
  \label{fig:arch}
\end{figure*}


As shown in Figure \ref{fig:arch}, our attention-guided multimodal SNN framework firstly processes heterogeneous modalities through specialized subnetworks to enhance the feature extraction. 
Following modality-specific feature extraction, time alignment modules (adapted from \cite{han2024balanced}) synchronize multi-modal spike trains to a unified temporal resolution. Cross-modal fusion is achieved through concatenation or weighted summation, producing joint representations that preserve both spatial and sequential patterns. Notably, a temporal attention-guided module then dynamically assigns adaptive temporal-wise weight to fused features at each timestep, where attention scores are learned by the temporal attention module through backpropagation to emphasize task-crucial temporal segments. These scores directly modulate temporal information integration in downstream spiking layers, enabling context-aware feature fusion during the final spike generation for classification. To optimize multi-modal synergy, the loss function incorporates dual mechanisms: (1) attention-driven temporal weighting adjusts the contribution of each timestep’s output to the total loss, and (2) a gradient modulation mechanism dynamically scales learning rates for individual modalities based on their evolving contributions, measured by their error propagation efficiency. Shared classifier parameters between unimodal and multimodal pathways ensure consistent feature representation while reducing model complexity. This architecture achieves adaptive spatiotemporal fusion and balanced multi-modal optimization without manual heuristic design.

\textbf{Modality-Specific Feature Extraction Branch.} In our multimodal SNN model, each modality pathway could employ specialized spiking neurons optimized for its temporal characteristics. In this paper, we employ the spike-based ResNet18 architecture for each modality branch on the CREMA-D and AVE datasets, followed by the concatenate feature fusion, the temporal attention module, and the final classification layer. In detail, the recurrent leaky integrate-and-fire (LIF) neuronal framework formulated through Euler discretization is employed to enhance temporal information aggregation and signal propagation during sequential processing \cite{deng2021temporal, shen2023esl, huang2025differential}. Based on the preceding membrane state at time step $t-1$ and accumulated presynaptic signals$I(t)$, the potential $u_j(t)$ for the postsynaptic unit $j$ evolves through:
\begin{equation}
    u_j(t) = \tau u(t-1) + I(t),
\end{equation}
where the leakage coefficient $\tau$ remains constant at $0.5$ throughout the simulation. The input current $I(t)$ emerges from the linear combination of connection weights $W$ and spike trains $x(t)$. Consistent with conventional spiking neurons analyzed previously, when $u(t)$ exceedsthe threshold $V_{th}$, the neuron generates an action potential followed by membrane reset to baseline. This mechanism produces binary neural activation $a(t+1)$ and potential reset governed by:
\begin{equation}
    a(t+1) = \Theta \left( u(t+1) - V_{th} \right),
\end{equation}
\begin{equation}
    u(t+1) = u(t+1) \left( 1 - a(t+1) \right).
\end{equation}

After the feature extractor in our multimodal SNNs, the convolutional time alignment module is employed to adjust the temporal scales of key features corresponding to different unimodal information, thereby determining the temporally aligned features. We take an example that contains visual and audio modalities (we also use this example in the following content in the method section); then the convolutional time alignment module comprises two components: a visual convolutional time alignment module and an auditory convolutional time alignment module. These modules are designed to align the crucial temporal features of visual and auditory modalities, respectively, ensuring synchronization across different time scales. This approach enables more accurate integration and analysis of multimodal data by resolving temporal discrepancies between modalities. Further details on the convolutional time alignment mechanism could refer to \cite{han2024balanced}. After the time step alignment by convolutional kernels, the spike-based features from each modality branch would begin to fuse.

\subsection{Temporal Attention-guided Adaptive Fusion  Module}

The TAAF module computes temporal importance scores for every timestep by modeling the relationships across timesteps in the fusion spike-based feature sequence. These scores are then integrated into the above TET loss to enable dynamic adjustment of loss weights for individual timesteps.

Meanwhile, we need to simulate forward propagation through matrix multiplication between time-aligned unimodal spike-based features and multimodal classifier parameters to obtain unimodal and multimodal outputs, then evaluate each modality's contribution ratio to determine modulation factors. 

In practical implementation, the process of obtaining multimodal outputs and calculating modality contributions through feedforward simulation comprises two parts. The first one is multimodal output computation: The classifier contains a 256-ClassNumber spiking fully connected layer. Let $W$ denote weight parameters and $b$ the bias term. The multimodal output $O_f^i$ is calculated as:
$O_f^i = W[h_{m1}^i, h_{m2}^i] + b$. The second one is unimodal feedforward simulation: To estimate modality-specific contributions, we decompose the original classifier into visual (128-ClassNumber) and auditory (128-ClassNumber) sub-classifiers with $W = [W_{m1}, W_{m2}]$. The unimodal outputs are computed as:
$O_{m1}^i = W_{m1} h_{m1}^i + b/2$, $O_{m2}^i = W_{m2} h_{m2}^i + b/2$. Thus we obtain the multimodal output and unimodal outputs as follows:

\begin{equation}
\centering
\begin{aligned}
 O_f &= F(h_{m1}, h_{m2}), \\
 O_u & = Classifier(h_u), u \in \{m1,m2\} ,
\end{aligned}
\end{equation}

where $h_{m1}, h_{m2}$ denotes two different unimodal spiking-based features (for instance, $h_{m1}, h_{m2}$ could be the visual image and audio modality in the CREMA-D dataset). $F(\cdot)$ denotes the multimodal fusion layer in multimodal SNNs, which fuses the unimodal features from each feature extractor of the modality-specific branch. Thus, $O_f$ represents fused feature outputs after feature fusion. Here $O_u$ contains $O_{m1}$ and $O_{m2}$ for different unimodal features.

Based on the above unimodal and fused features in our multimodal SNNs, the TAAF  modules compute temporal importance scores that capture cross-modal dependencies across timesteps:

\begin{equation}
\centering
\begin{aligned}
& Q_u = O_uW^Q_u, \\
&  K_u = O_uW^K_u, 
\end{aligned}
\end{equation}
where $W^Q_u, W^K_u \in \mathbb{R}^{D \times C}, \quad u \in \{m1,m2,f\}$. The learnable parameter matrices $W^Q_u, W^K_u$ project the feature outputs of each modality into query and key representations, followed by scaled dot-product operations to compute inter-modal similarity matrices across temporal dimensions (different time steps in our multimodal SNNs). Based on that, cross-timestep dependencies can be captured by modeling dynamic correlations between heterogeneous modalities at varying temporal resolutions. That is, the temporal similarity matrix is computed as:

\begin{equation}
\centering
\begin{aligned}
& A(Q_u, K_u) = \text{softmax}\left(\frac{Q_uK_u^T}{\sqrt{C}}, \text{dim}=-1\right),
\end{aligned}
\end{equation}

Next, the temporal attention score $\alpha_u$ can be obtained by averaging the temporal similarity matrix:

\begin{equation}
\label{eqattentionscore}
\centering
\begin{aligned}
& \alpha_u = \frac{1}{T}\sum_{i=1}^T A_{i}(Q_u, K_u).
\end{aligned}
\end{equation}

\subsection{Temporal Adaptive Balanced Fusion Loss}

Based on the above TAAF module, our novel loss function adaptively weights temporal features 
 in each time step through attention guidance:

\begin{equation}
\label{eqlossagl}
\mathcal{L}_{\text{AGLori}} = -\sum_{t=1}^T \alpha_u(t) \sum_{i=1}^n \hat{y}_i \log S(O^i_u(t)),
\end{equation}
where $S(\cdot)$ denotes softmax normalization and $\alpha_u(t)$ modulates loss contributions at each timestep. $\hat{y}_i$ denotes the one-hot code of the target label. $O^i_u(t)$ denotes the output of modality 
$u$ at timestep $t$ for $i_{th}$ sample.
We can compute the losses for unimodal and multimodal outputs based on $\mathcal{L}_{\text{AGL}}$, then dynamically adjust the final loss using modulation factors to optimize multimodal objectives. Here we represent the losses of two different unimodals as $\mathcal{L}_{m1}$ and $\mathcal{L}_{m2}$, and the fused multimodal as $\mathcal{L}_f$.
Furthermore, to improve the robustness of the above loss function, the normalization of mean square error $\mathcal{L}_{\text{MSE}} = 1/T \sum_{t=1}^{T} MSE(O(t), \Phi)$ is introduced in the final $\mathcal{L}_{\text{AGL}}$, where $\Phi$ is the hyperparameter of the membrane potential normalization. Therefore, the final $\mathcal{L}_{\text{AGL}}$ is :
\begin{equation}
\label{eqlossagl}
\mathcal{L}_{\text{AGL}} = (1-\lambda) \mathcal{L}_{\text{AGLori}} + \lambda \mathcal{L}_{\text{MSE}}.
\end{equation}

Meanwhile, here to obtain the final loss function for balanced multimodal SNN learning, we compute the modality contribution ratio by processing unimodal outputs through softmax to obtain modality scores:$s_{m1} = \sum_{j=1}^C \mathbf{1}_{\{j=y_i\}} \text{softmax}(O_{m1}^i)_j$ and 
$s_{m2} = \sum_{j=1}^C \mathbf{1}_{\{j=y_i\}}$ $ \text{softmax}(O_{m2}^i)_j$,
where $C$ denotes class number, $y_i$ the true label. Then the batch-level modality contribution ratios are computed as:
$\rho_{m1}^t = \sum_{i \in B_t} s_{m1}^i  / \sum_{i \in B_t} s_{m2}^i$, and
$\rho_{m2}^t = \sum_{i \in B_t} s_{m2}^i  $ $/ \sum_{i \in B_t} s_{m1}^i$.
Hence, the modality-specific factors $k_u^t$ ($u \in \{m1,m2\}$) are calculated via: $k_u^t = 1 - \tanh(\alpha_u \rho_u^t)$ if $\rho_u^t > 1 $, otherwise $k_u^t$ equals to 1, where $\alpha_u$ controls suppression intensity for dominant modalities.

Thus, the final loss integrates unimodal and fused multimodal learning for our proposed multimodal SNNs:
\begin{equation}
\label{eqfinalloss}
\mathcal{L} = \beta \times (k_{m1}\mathcal{L}_{m1}+ k_{m2}\mathcal{L}_{m2}) + \mathcal{L}_f,
\end{equation}
where $\beta$ controls modality competition balance  $k_{m1}/k_{m2}$ are attention-guided scaling factors.
Combined with Algorithm \ref{alg:training}, the proposed multimodal SNNs can be trained efficiently.

\begin{algorithm}[t]
\caption{Attention-Guided Multimodal SNN Training}
\label{alg:training}
\begin{algorithmic}[1]
\STATE Initialize SNN parameters $\theta$, attention matrices $W^Q_u, W^K_u$
\FOR{epoch $= 1$ to $N$}
  \FOR{each batch $(X_{m1}, X_{m2}, y)$}
    \STATE Feature extractor forward propagates through SNN branches for each modality to get $h_{m1}, h_{m2}$.
    \STATE Fuse modalities: $O_f = F(h_{m1}, h_{m2})$
    \STATE Compute unimodal outputs: $O_{m1}, O_{m2} $
    \STATE Calculate temporal attention scores $\alpha_u$ via Eq. \ref{eqattentionscore}
    \STATE Compute losses $\mathcal{L}_{m2}, \mathcal{L}_{m2}, \mathcal{L}_f$ using Eq. \ref{eqlossagl}
    \STATE Update $\mathcal{L}$ with neuromodulatory fusion (Eq. \ref{eqfinalloss})
    \STATE Backpropagate errors and update parameters $\theta$
  \ENDFOR
  \STATE Adjust modality weights $k_{m1}, k_{m2}$ based on validation performance
\ENDFOR
\end{algorithmic}
\end{algorithm}

\begin{table*}
\caption{Comparative study of different methods (contains ANN and SNN methods) with various fusion strategies on AVE and CREMA-D datasets.}
\begin{tabular}{cccccc}
\hline
Dataset                                               & Model                                     & Method                                                                                     & Fusion Strategy        & Time Steps    & Accuracy (\%)  \\ \hline
\multicolumn{1}{c|}{\multirow{17}{*}{CREMA-D  dataset}} & \multicolumn{1}{c|}{\multirow{9}{*}{ANN}} & \multirow{4}{*}{\begin{tabular}[c]{@{}c@{}}OGM-GE \cite{peng_balanced_2022}\end{tabular}} & Concatenation          & /             & 61.90          \\
\multicolumn{1}{c|}{}                                 & \multicolumn{1}{c|}{}                     &                                                                                            & Summation              & /             & 62.20          \\
\multicolumn{1}{c|}{}                                 & \multicolumn{1}{c|}{}                     &                                                                                            & Film                   & /             & 55.60          \\
\multicolumn{1}{c|}{}                                 & \multicolumn{1}{c|}{}                     &                                                                                            & Gated                  & /             & 60.62          \\ \cline{3-6} 
\multicolumn{1}{c|}{}                                 & \multicolumn{1}{c|}{}                     & \multirow{4}{*}{PMR \cite{fan2023pmr}}                                                     & Concatenation          & /             & 61.10          \\
\multicolumn{1}{c|}{}                                 & \multicolumn{1}{c|}{}                     &                                                                                            & Summation              & /             & 59.40          \\
\multicolumn{1}{c|}{}                                 & \multicolumn{1}{c|}{}                     &                                                                                            & Film                   & /             & 61.80          \\
\multicolumn{1}{c|}{}                                 & \multicolumn{1}{c|}{}                     &                                                                                            & Gated                  & /             & 59.90          \\ \cline{3-6} 
\multicolumn{1}{c|}{}                                 & \multicolumn{1}{c|}{}                     & AGM \cite{li2023boosting}                                                                  & /                      & /             & 70.16          \\ \cline{2-6} 
\multicolumn{1}{c|}{}                                 & \multicolumn{1}{c|}{\multirow{8}{*}{SNN}} & WeightAttention \cite{liu2022event}                                                        & /                      & 4             & 64.78          \\ \cline{3-6} 
\multicolumn{1}{c|}{}                                 & \multicolumn{1}{c|}{}                     & SCA \cite{guo2023transformer}                                                              & /                      & 4             & 66.53          \\ \cline{3-6} 
\multicolumn{1}{c|}{}                                 & \multicolumn{1}{c|}{}                     & CMCI \cite{zhou2024enhancing}                                                              & /                      & 4             & 70.02          \\ \cline{3-6} 
\multicolumn{1}{c|}{}                                 & \multicolumn{1}{c|}{}                     & S-CMRL \cite{he2025enhancing}                                                              & /                      & 4             & 73.25          \\ \cline{3-6} 
\multicolumn{1}{c|}{}                                 & \multicolumn{1}{c|}{}                     & MISNET-L \cite{liutowards}                                                                 & Concatenation          & Various       & 75.22          \\
\multicolumn{1}{c|}{}                                 & \multicolumn{1}{c|}{}                     & MISNET-XL \cite{liutowards}                                                                & Concatenation          & Various       & 77.14          \\ \cline{3-6} 
\multicolumn{1}{c|}{}                                 & \multicolumn{1}{c|}{}                     & \multirow{2}{*}{\textbf{TAAF-SNNs (Ours)}}                                       & \textbf{Concatenation} & \textbf{3}    & \textbf{77.55} \\
\multicolumn{1}{c|}{}                                 & \multicolumn{1}{c|}{}                     &                                                                                            & \textbf{Summation}     & \textbf{3}    & \textbf{76.75} \\ \hline
\multicolumn{1}{c|}{\multirow{12}{*}{AVE dataset}}    & \multicolumn{1}{c|}{\multirow{8}{*}{ANN}} & \multirow{4}{*}{OGM-GE \cite{peng_balanced_2022}}                                          & Concatenation          & /             & 65.42          \\
\multicolumn{1}{c|}{}                                 & \multicolumn{1}{c|}{}                     &                                                                                            & Summation              & /             & 66.42          \\
\multicolumn{1}{c|}{}                                 & \multicolumn{1}{c|}{}                     &                                                                                            & Film                   & /             & 64.43          \\
\multicolumn{1}{c|}{}                                 & \multicolumn{1}{c|}{}                     &                                                                                            & Gated                  & /             & 62.69          \\ \cline{3-6} 
\multicolumn{1}{c|}{}                                 & \multicolumn{1}{c|}{}                     & \multirow{4}{*}{PMR \cite{fan2023pmr}}                                                     & Concatenation          & /             & 67.10          \\
\multicolumn{1}{c|}{}                                 & \multicolumn{1}{c|}{}                     &                                                                                            & Summation              & /             & 68.10          \\
\multicolumn{1}{c|}{}                                 & \multicolumn{1}{c|}{}                     &                                                                                            & Film                   & /             & 66.40          \\
\multicolumn{1}{c|}{}                                 & \multicolumn{1}{c|}{}                     &                                                                                            & Gated                  & /             & 62.70          \\ \cline{2-6} 
\multicolumn{1}{c|}{}                                 & \multicolumn{1}{c|}{\multirow{4}{*}{SNN}} & MISNET-XL \cite{liutowards}                                                                & Concatenation          & Various       & 67.24          \\
\multicolumn{1}{c|}{}                                 & \multicolumn{1}{c|}{}                     & MISNET-XL \cite{liutowards}                                                                & Concatenation          & Various       & 68.04          \\ \cline{3-6} 
\multicolumn{1}{c|}{}                                 & \multicolumn{1}{c|}{}                     & \multirow{2}{*}{\textbf{TAAF-SNNs (Ours)}}                                       & \textbf{Concatenation} & \textbf{4--3} & \textbf{70.65} \\
\multicolumn{1}{c|}{}                                 & \multicolumn{1}{c|}{}                     &                                                                                            & \textbf{Summation}     & \textbf{4--3} & \textbf{68.91} \\ \hline
\end{tabular}
\label{ComparisonAVECreamd}
\end{table*}

\section{Results}

We systematically evaluate our framework through three experimental phases: comparative performance analysis among recent ANN and SNN multimodal learning methods, ablation studies validating the efficacy of our attention-guided training mechanism, and convergence dynamics analysis demonstrating its capability to mitigate modality imbalance. First, we benchmark our approach against state-of-the-art multimodal models on two datasets. Subsequently, ablation experiments isolate the contributions of temporal lossadaptation and attention-guided fusion. Finally, epoch-wise analyses of accuracy and loss trajectories for unimodal and multimodal training reveal how our method harmonizes cross-modal learning dynamics.

The evaluation mainly employs three multimodal benchmarks. The first one is CREMA-D \cite{cao2014crema}, an audio-visual speech emotion recognition dataset containing 7,442 video clips labeled with six emotional categories (happy, sad, anger, fear, disgust, and, neutral). The dataset is partitioned into 6,698 training and 744 testing samples, with each clip capturing synchronized facial expressions and vocal utterances. 
Another benchmark for audio-visual event localization and classification is AVE \cite{tian2018audio}, which comprises 4,143 ten-second videos across 28 event categories (e.g., "man speaking," "dog barking," "playing guitar"). Each video contains temporally aligned auditory and visual events, challenging models to correlate cross-modal cues at fine temporal resolutions. 
The final benchmark is the Event-Audio Digit Recognition (EAD) datasets. The event-based data contains discrete visual events which can be applied into different computer vision tasks \cite{liu2024line, liu2024optical, liu2025stereo, liu2024lecalib}.  The network architecture employs the convolutional SNNs and recurrent SNNs for the event and audio modalities followed by \cite{han2024balanced}.  

\subsection{Performance Comparison}

In this section, we conduct the performance comparison among different multimodal ANNs and SNNs methods on AVE and CREMA-D datasets. The compared ANNs methods contain 
OGM-GE \cite{peng_balanced_2022}, PMR \cite{fan2023pmr}, and AGM \cite{li2023boosting}. While the compared multimodal SNNs contain WeightAttention \cite{liu2022event}, SCA \cite{guo2023transformer}, CMCI \cite{zhou2024enhancing}, and S-CMRL \cite{he2025enhancing}. Especially, the last two compared SNNs models employ the spike-based Transformer as the backbone, and their performance is relatively advanced in the multimodal SNNs models.

As shown in Table \ref{ComparisonAVECreamd},  our proposed Attention-Guided SNN demonstrates state-of-the-art performance on both CREMA-D and AVE datasets, achieving 77.55\% and 70.65\% accuracy respectively, while operating at significantly fewer timesteps (3-4) compared to competing SNN methods.
In detail, despite using fewer timesteps than conventional SNNs (3 vs. 4-20+ in MISNET variants), our model outperforms all baselines on CREMA-D. This aligns with our hypothesis that time-variant attention scoring enables precise extraction of spatiotemporal features—where crucial modality-specific spike patterns are amplified at optimal phases (e.g., visual motion cues at early timesteps, auditory phonemes at later ones). In contrast, fixed-timestep SNNs like CMCI (70.02\% at T=4) waste computational resources propagating redundant spikes, as evidenced by their 7.53\% lower accuracy than our method despite comparable timesteps. These advanced performances of our model suggest the importance of temporal efficiency via dynamic temporal attention modulation.
Meanwhile, the superior performance over ANN baselines (70.16\% AGM vs. our 77.55\% on CREMA-D) stems from our biologically grounded fusion strategy. While ANNs rely on static concatenation/summation (61.90–70.16\%), our attention-guided fusion brings more abundant temporal dynamics by mimicking cortical multisensory integration principles. Especially, our proposed temporal attention scores in each time step of SNNs enhance the cross-modal interactions.
Moreover, the minimal performance gap between concatenation (77.55\%) and summation (76.75\%) fusion in our model contrasts sharply with ANN counterparts (e.g., OGM-GE: 61.90\% vs. 62.20\%). This demonstrates that temporal attention scores decouple fusion effectiveness from architectural constraints—a crucial advantage for neuromorphic hardware implementations where concatenation is often infeasible.
In addition, we obtain the accuracy of 97.59\% on the EAD dataset under 5 time steps without time step alignment for both two modalities. Comparing with the WeightAttention \cite{liu2022event} (96.45\%) and OGM-GE \cite{peng_balanced_2022} (95.3\%), our model improves the fusion performance further by considering the temporal dynamics for each time step of SNNs.

\begin{table*}
\centering
\caption{Ablation Study on the proposed temporal attention-guided module.}
\begin{tabular}{cccc}
\hline
Method                                                                                                         & Fusion Strategy & AVE            & CREMA-D          \\ \hline
\multicolumn{1}{c|}{\multirow{4}{*}{SNN with LA}}                                                              & Concatenation   & 69.65          & 75.27          \\
\multicolumn{1}{c|}{}                                                                                          & Summation       & 69.41          & 74.60          \\
\multicolumn{1}{c|}{}                                                                                          & Film            & 66.41          & 65.72          \\
\multicolumn{1}{c|}{}                                                                                          & Gated           & 63.68          & 68.01          \\ \hline
\multicolumn{1}{c|}{\multirow{4}{*}{\begin{tabular}[c]{@{}c@{}}SNN with LA \\ and TA\end{tabular}}}            & Concatenation   & 69.90          & 75.13          \\
\multicolumn{1}{c|}{}                                                                                          & Summation       & 69.15          & 75.67          \\
\multicolumn{1}{c|}{}                                                                                          & Film            & 64.68          & 66.67          \\
\multicolumn{1}{c|}{}                                                                                          & Gated           & 67.66          & 68.82          \\ \hline
\multicolumn{1}{c|}{\multirow{2}{*}{\begin{tabular}[c]{@{}c@{}}TAAF-SNNs\\  without TA\end{tabular}}} & Concatenation   & \textbf{69.90} & \textbf{76.75} \\
\multicolumn{1}{c|}{}                                                                                          & Summation       & \textbf{69.15} & \textbf{79.17} \\ \hline
\multicolumn{1}{c|}{\multirow{2}{*}{\begin{tabular}[c]{@{}c@{}}TAAF-SNNs \\ (ours) \end{tabular}}}           & Concatenation   & \textbf{70.65} & \textbf{77.55} \\
\multicolumn{1}{c|}{}                                                                                          & Summation       & \textbf{68.91} & \textbf{76.75} \\ \hline
\end{tabular}
\label{tableAblationstudy}
\end{table*}

\begin{figure*}[]
    \centering
        \begin{tabular}{cccccc}
        \includegraphics[width=1\textwidth]{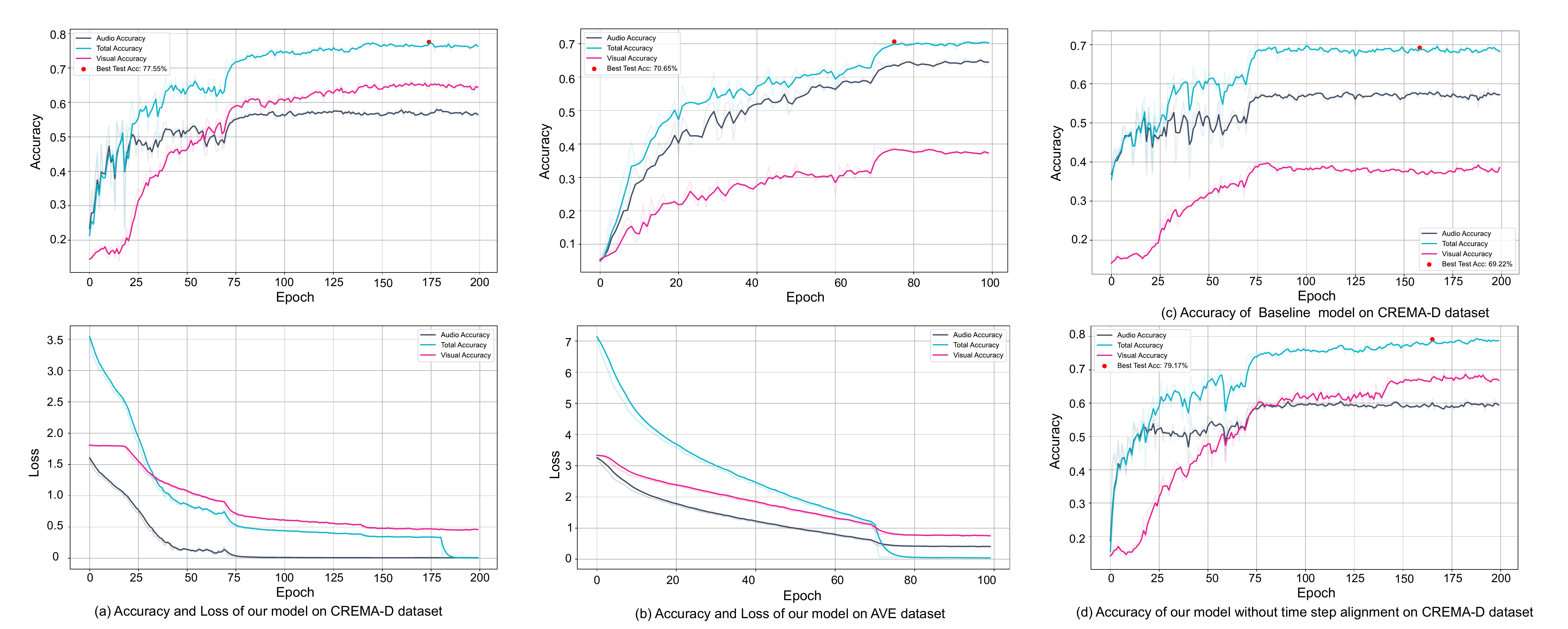} 
        \end{tabular}

    \caption{The epoch-wise analysis of accuracy and loss trajectories with the proposed multimodal SNNs on CREMA-D and AVE datasets. (a) The top and bottom figures are the accuracy and loss trajectories for CREMA-D datasets, respectively. (b) The top and bottom figures are the accuracy and loss trajectories for AVE datasets, respectively. (c) The accuracy trajectory of the baseline model without the proposed TAAF. (4) The accuracy trajectory of our model without time step alignment module. }
    \label{BestAccCreamdAVE}
\end{figure*}

\subsection{Ablation Study}

As shown in Table \ref{tableAblationstudy}, the ablation study validates the crucial role of our temporal attention-guided mechanism and its synergy with loss adjustment among different modalities, highlighting the necessity of each component in addressing modality imbalance. LA and TA mean the loss adjustment method without TAAF and time step alignment, respectively.
Our full temporal attention-guided multimodal SNN achieves superior performance across fusion strategies, with 70.65\% (AVE) and 77.55\% (CREMA-D) for concatenation—surpassing loss adjustment-only variants by 1.0–3.2\%. Crucially, the performance gap between concatenation and summation shrinks to <1.7\% (vs. 3.5–5.3\% in only loss adjustment models), demonstrating that time step alignment decouples fusion quality from architectural constraints. This aligns with our emphasis on temporal dynamic prioritization over static fusion heuristics.
Moreover, the attention-guided mechanism critically enhances fusion by dynamically reweighting temporal contributions of differenttime steps. 
In the full model, concatenation fusion exhibited high attention score variance, indicating focus on crucial timesteps. This drove a 1.8\% accuracy gain on CREMA-D (77.55\% vs. 75.67\% without attention).

\subsection{Modality Imbalance Analysis}

As shown in Figure \ref{BestAccCreamdAVE} (a) and (b), our epoch-wise analysis of accuracy  and loss trajectories reveals crucial insights into modality competition and the effectiveness of our attention-guided temporal loss adaptation. The results across the CREMA-D and AVE datasets demonstrate how our method resolves modality imbalance through time-step-aware modulation of cross-modal dynamics.
In the early training phase (Epochs 1–55) on the CREMA-D dataset, the audio modality dominates the model training, which shows rapid convergence (with accuracies from 20\% to 50\%) due to its temporally discriminative features in emotion recognition.
In this phase, the visual modality lags behind the audio modality. Then the gradient modulation factor gradually enhances visual updates (with the accuracy from 10\% to 50\%) to prevent premature dominance. At about epoch 55, visual accuracy surpasses audio as attention-guided loss prioritizes underlearned modality. After a while, the cross-modal accuracy jumps from about 63\% to over 70\% at epoch 75, leveraging complementary strengths (audio’s prosody and visual expressions) until the final convergence. 
Meanwhile, on the AVE dataset, the audio branch accuracy consistently leads during the whole training process, which reflects its crucial role in event localization.  Despite audio’s advantage, fusion accuracy still exceeds both modalities, proving synergistic integration because of the proposed attention-guided multimodal SNNs learning.
In addition, the loss trajectory of the multimodal model appears to exhibit a sharper decreasing phenomenon than these two single-modality loss trajectories. Also, the decreasing speed of loss convergence of multimodal becomes quite sharper at epoch 180, which is because the modulation of the learning speed for each branch of each modality is ended by the hyperparameter settings. This is to guarantee the final multimodal convergence performance after the modulation of modality imbalance. This phenomenon appears more obviously on the AVE dataset (shown in Figure \ref{BestAccCreamdAVE} (b)), where the modulation end epoch is set to be 70. 
In addition, we visualize the accuracy trajectory of the baseline model without the proposed TAAF module as shown in Figure \ref{BestAccCreamdAVE} (c). Without the TAAF module with temporal adaptive balanced fusion loss, the modality dominance remains unchanged during the whole training process, which limits the final multimodal performance due to the ignorance of fusion feature adjustment.
In contrast, we also provide the performance visualization of our model without time step alignment under the summation feature fusion strategy. Since the time step alignment module after the feature extractor modules mainly aims to decrease the time step redundancy, we infer that the proposed TAAF could enhance the feature fusion in the temporal dimension, especially under the summation feature fusion strategy.
Therefore, the proposed model could implement that the attention weights shift focus between modalities based on their temporal learning progress. Meanwhile, the fusion loss convergence outpaces unimodal training, which suggests the efficient cross-modal credit assignment through our model. Also, sustained audio dominance (in AVE) could also coexist with balanced fusion, as the mechanism preserves modality-specific strengths while optimizing joint representations.
These results confirm that temporally granular loss modulation—not merely modality balancing—is key to unlocking SNNs’ multimodal potential, achieving state-of-the-art performance while maintaining biological plausibility.

In addition, we conduct the energy consumption estimation on the CREMA-D dataset. The estimated energy consumption of the proposed TAAF-SNNs and the ANNs with the similar architecture (with about 25M parameters) is 18.5 MJ and 49.91 MJ, respectively. That emphasizes the energy efficiency advantage of TAAF-SNNs.

\section{Conclusion}

This paper addresses the dual challenges of modality imbalance and temporal misalignment in multimodal SNNs through biologically plausible temporal attention mechanisms. The proposed TAAF module improves the temporal hierarchy by prioritizing modality-specific processing and dynamic fusion feature weighting in each time step. By integrating temporal attention scores into both feature fusion and loss calculation, our framework achieves higher accuracy than conventional spike averaging methods while balancing modality convergence. Further, the temporal adaptive balanced fusion loss implements the gradient modulation and enables autonomous compensation for sensory processing asymmetries, a crucial advancement for real-world scenarios where modalities exhibit inherent temporal disparities.
Our experiments demonstrate that modeling temporal dynamics at each time step for SNNs, cross-modal attention-guided fusion, and loss-based gradient balancing can yield substantial improvements in both accuracy and energy efficiency. The framework has the potential to implement deployment with neuromorphic hardware for multisensory classification tasks. Future work will extend these principles to spiking transformer architectures and explore applications in neurorobotics, where temporal coherence across sensory modalities is crucial for embodied intelligence. This research advances multimodal SNNs toward biological fidelity while establishing practical benchmarks for adaptive temporal processing in energy-constrained AI systems.


\section{Acknowledgements}

This work was supported by National Natural
Science Foundation of China under Grant (No.
62306274, 62088102, 62476035, 62206037, 61925603), Open Research Program of the National Key Laboratory of Brain-Machine Intelligence, Zhejiang University (No. BMI2400012).

\bibliographystyle{unsrt}
\bibliography{main}




\end{document}